\documentclass[conference]{IEEEtran}
\makeatletter
\IEEEoverridecommandlockouts
\usepackage{cite}
\usepackage{amsmath,amssymb,amsfonts}
\usepackage{algorithmic}
\usepackage{graphicx}
\usepackage{url}             
\usepackage{textcomp}
\usepackage{xcolor}
\usepackage{multirow} 
\usepackage{booktabs}
\usepackage{diagbox}
\usepackage{makecell} 
\usepackage{booktabs} 

\def\BibTeX{{\rm B\kern-.05em{\sc i\kern-.025em b}\kern-.08em
    T\kern-.1667em\lower.7ex\hbox{E}\kern-.125emX}}
\begin{document}
\vspace{-30pt} 

\title{Patch-Wise Hypergraph Contrastive Learning with Dual Normal Distribution Weighting for Multi-Domain Stain Transfer

}
\vspace{-40pt} 
\author{
\IEEEauthorblockN{Haiyan Wei\textsuperscript{1}, Hangrui Xu\textsuperscript{1}, Bingxu Zhu\textsuperscript{1},\\ 
Yulian Geng\textsuperscript{1}, Aolei Liu\textsuperscript{1}, Wenfei Yin\textsuperscript{1}, Jian Liu\textsuperscript{1\textdagger}}
\IEEEauthorblockA{
\textsuperscript{1}Hefei University of Technology, Hefei, China \\
Emails: 2022217594@mail.hfut.edu.cn, 2022217415@mail.hfut.edu.cn, \\
2022217593@mail.hfut.edu.cn, gengyl626@mail.hfut.edu.cn, \\
lal@mail.hfut.edu.cn, wenfeiyin@hfut.edu.cn, \textsuperscript{\textdagger}jianliu@hfut.edu.cn
}
}

\maketitle
\vspace{-10pt}
\footnotetext[1]{\textdagger Corresponding author}

\vspace{-18pt} 
\maketitle
\vspace{-18pt} 
\begin{abstract}

Virtual stain transfer leverages computer-assisted technology to transform the histochemical staining patterns of tissue samples into other staining types. However, existing methods often lose detailed pathological information due to the limitations of the cycle consistency assumption. To address this challenge, we propose STNHCL, a hypergraph-based patch-wise contrastive learning method. STNHCL captures higher-order relationships among patches through hypergraph modeling, ensuring consistent higher-order topology between input and output images. Additionally, we introduce a novel negative sample weighting strategy that leverages discriminator heatmaps to apply different weights based on the Gaussian distribution for tissue and background, thereby enhancing traditional weighting methods. Experiments demonstrate that STNHCL achieves state-of-the-art performance in the two main categories of stain transfer tasks. Furthermore, our model also performs excellently in downstream
tasks. Code is available at \url{https://github.com/Whywwwzzzg/STNHCL}.
\end{abstract}

\vspace{-18pt} 
\section{Introduction}
\label{sec:intro}

Whole Slide Images (WSIs) are the gold standard in histopathology for clinical diagnosis, utilizing stains or fluorescent markers to visualize tissue structures. Hematoxylin and eosin (H\&E) staining is favored for its cost-effectiveness. Other methods like masson’s trichrome (MAS), periodic acid-schiff (PAS) and acid-silver-methenamine (PASM) target specific features such as collagen fibers and glycoproteins. Immunohistochemical (IHC) emphasizes specific epitopes via antigen-antibody binding \cite{bai2023deep}. However, compared to H\&E, these other stains often require more labor-intensive and costly tissue processing. Deep learning-based virtual stain transfer reduces labor and costs of additional staining and imaging \cite{bai2023deep},\cite{lo2021cycle}. For many diseases, multiple staining types are needed for better diagnostic information. Recent works \cite{lin2022unpaired},\cite{guan2024unsupervised} have achieved one-to-many stain transfer with a single network, generating multiple target stains from H\&E images. 
\begin{figure}[htbp]
\centerline{\includegraphics[scale=0.045]{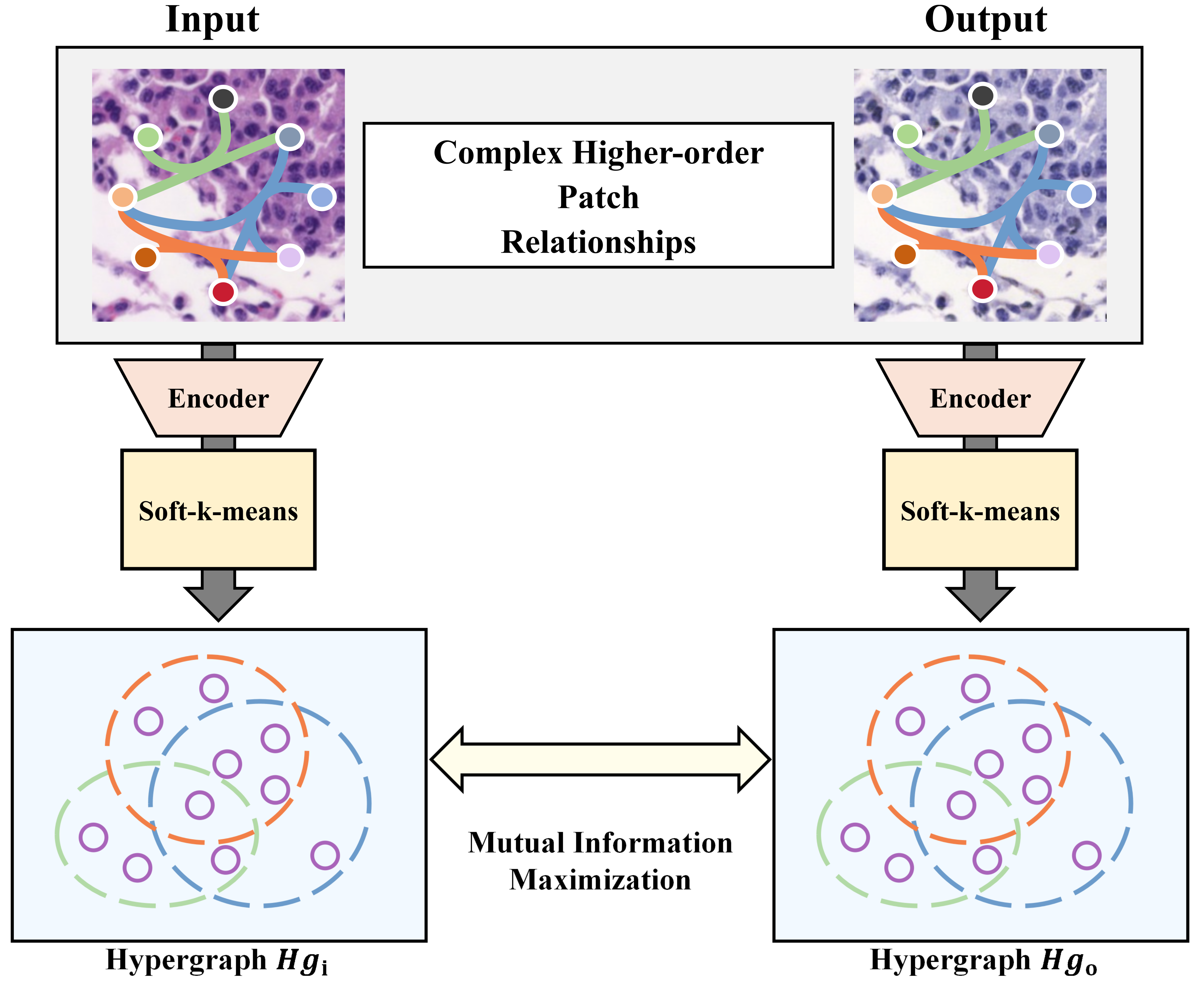}} 
\caption{There are implicit higher-order semantic connections between the input and output images. We model these higher-order semantics using a hypergraph and maximize the mutual information between them.}
\label{fig1}
\vspace{-20pt}
\end{figure}

However, these methods rely on cycle-consistency loss\cite{zhu2017unpaired} to maintain consistency between source and generated images. This assumption often limits style variation in complex restaining tasks and neglects critical semantic details, leading to loss of important pathological information. Recently, inspired by the success of contrastive learning strategies, patch-wise contrastive learning \cite{park2020contrastive} has been introduced into image translation. By maximizing mutual information across corresponding patches in the input and output images, patch-wise contrastive learning improves the modeling capability for content consistency. Some methods have refined patch-wise contrastive learning to enhance the quality of generated images \cite{zhan2022modulated},\cite{hu2022qs}. Among them, PatchGCL\cite{jung2024patch} constructs a graph of patch features and enforces topological consistency between the input and output image features to preserve high-level semantic content. However, using simple graphs to represent images has its limitations\cite{han2023vision}. In a simple graph structure, each edge connects only two nodes, which inherently restricts it to capturing pairwise relationships. In patch-wise image translation tasks, the relationships between image patches are often far more complex, involving higher-order semantic correlations that go beyond simple pairwise interactions. These higher-order correlations are critical for accurately modeling the complex semantic relationships between patches in the input and output images. Moreover, graph-based representations of patch features often lead to redundant associations. Specifically, when multiple adjacent patches share similar features, the graph construction process tends to generate excessive and redundant edges, resulting in an inefficient and less expressive representation of patch dependencies. To address these issues, we propose STNHCL, which employs hypergraph modeling of image features. A hypergraph is a generalized graph structure. Unlike simple graphs that only connect pairs of nodes, hypergraphs can have hyperedges connecting any number of nodes. Hypergraphs excel at capturing complex higher-order correlations present in images\cite{han2023vision}. We treat patches in patch-wise contrastive learning as hypergraph nodes and use soft k-means to build connections, maintaining higher-order topological consistency between input and output images, as shown in Fig. \ref{fig1}. 
Furthermore, to enhance the utilization of negative samples, we propose a novel approach that integrates both the hard weighting strategy for prior unpaired image translation and the easy weighting strategy for paired image translation into a unified framework. Our contributions are summarized as follows:


\vspace{-2pt} 
\begin{itemize}

 \item 
We propose patch-wise hypergraph contrastive learning to address shortcomings of previous methods in high-order pathological semantic consistency. This is the first application of patch-wise contrastive learning in multi-domain stain transfer tasks. We innovatively model patch features using hypergraphs, enhancing the consistency of higher-order topological structures.
  \item We propose a unique negative sample weighting strategy. To mitigate the extremeness of traditional weighting methods, we use a discriminator output heatmap to divide stained images into tissue and background regions. We then apply distinct, milder normal distribution weights to negative samples to make better use of their information.
  \item We evaluated our method on human kidney dataset and human lung lesion dataset for H\&E to special stain and H\&E to IHC tasks, respectively. The experiments demonstrate that our model achieves state-of-the-art performance in both types of tasks.

\end{itemize}
\vspace{-3pt} 
\section{Related Work}
\vspace{-2pt} 
\subsection{Contrastive Learning in Unpaired Image Translation}\label{AA}
\vspace{-2pt} 
Early unpaired image-to-image translation (I2I) methods rely on cycle-consistency assumptions\cite{zhu2017unpaired}, which impose overly restrictive bijection constraints. CUT \cite{park2020contrastive} introduced contrastive learning to I2I to move beyond cycle-consistency, significantly improving translation quality. QS-Attn\cite{hu2022qs} selects key anchors through a query-selection attention module. MoNCE\cite{zhan2022modulated} adjusts pushing forces based on negative-anchor similarity and introduces optimal transport. PatchGCL\cite{jung2024patch} uses graph neural networks to explore topological structures at the patch level to enforce high-level semantic consistency.
\vspace{-2pt}
\subsection{Multi-Domain Stain Transfer}\label{BB}
One-to-one stain transfer methods demand high resources for multiple stains, leading to the rise of multi-domain stain transfer methods. UMDST \cite{lin2022unpaired} employs style-guided normalization to dynamically control the transfer direction. PPHM-GAN \cite{kawai2024virtual} achieves virtual transformation between multiple high-resolution stained images. GramGAN \cite{guan2024unsupervised} uses style encoding dictionary for progressive multi-domain stain transfer.

\subsection{Hypergraphs in Computer Vision}\label{CC}
High-order interactions are ubiquitous in complex systems and applications, including images. For example, hypergraphs can be used to model high-order correlations among 3D objects \cite{gao20123} for retrieval tasks. Hypergraph learning was applied to video object segmentation \cite{huang2009video}. Hypergraph neural networks have also been applied to action recognition \cite{zhou2022hypergraph} and multi-human mesh recovery \cite{huang2023reconstructing}. However, the application of hypergraph in low-level vision tasks is still not well-explored.

\section{METHOD} 
\subsection{Patch-Wise Hypergraph Contrastive Learning} 
In our model, the input stained image $ X $, after passing through the encoder $ G_{enc} $, generates feature maps at $ L $ intermediate layers, forming a feature stack $ \{Z_l\}^L $, where $ l \in \{1, 2, \ldots, L\} $ denotes the layer number, and $ Z_l = G_{enc}^l(X) \in \mathbb{R}^{c \times h \times w} $, with $ c $, $ h $, and $ w $ representing the number of channels, height, and width of the feature map, respectively. Here, we sample $ K $ patch features to obtain $ Z_l^k \in \mathbb{R}^c $, where $ k \in \{1, 2, \ldots, K\} $, and $ Z_l^k $ represents the feature vector of the $ k $-th patch in the $ l $-th layer of the encoder for the input image. 
Similarly, the generated output stained image $ Y $, processed by the encoder $ G_{enc} $, produces feature maps in its $ L $ layers, forming a feature stack $ \{V_l\}^L $. Again, we sample $ K $ patch features $ V_l^k \in \mathbb{R}^c $, where $ k \in \{1, 2, \ldots, K\} $, and $ V_l^k $ represents the embedding vector of the $ k $-th patch in the $ l $-th feature map for the output image $ Y $. 

We detail the process for the input stained image branch, with the output branch following a similar structure. We designate the sampled $ K $ patch features as the nodes of the input branch hypergraph $ H_{gi} $. A hypergraph can be defined as $ H_g = (V, E) $, where $ V $ is a set of $ K $ unique nodes, and $ E $ is a set of $ M $ hyperedges. Unlike ordinary graphs where edges connect two nodes, in hypergraphs a single hyperedge can connect multiple nodes, allowing hypergraphs to represent complex relationships more flexibly. The structure of a hypergraph can be described by a hyperedge matrix $ H_i \in \mathbb{R}^{M \times K} $:
\begin{equation}
H_{ij} = 
\begin{cases} 
k, & \text{if node } v_k \text{ belongs to hyperedge } e_i \\ 
-1, & \text{otherwise} 
\end{cases}
\end{equation}

The set of nodes connected to any hyperedge $ e_i $ is denoted as $ \mathcal{V}(e_i) = \{ v_j \in \mathcal{V} | H_{ij} \neq -1 \} $. The set of hyperedges connected to any node $ v_j $ is defined as $ \mathcal{E}(v_j) = \{ e_i \in \mathcal{E} | H_{ij} \neq -1 \} $. The degree of a node $ d(v_j) = |\mathcal{E}(v_j)| $ is defined as the number of hyperedges connected to it, while the degree of a hyperedge $ d(e_i) = |\mathcal{V}(e_i)| $ is the number of nodes it connects. Additionally, $ D_v $ and $ D_e $ denote the diagonal matrices of hyperedge degrees and node degrees, respectively.

\vspace{-2pt}
\begin{figure*}[htbp]
\centering
\centerline{\includegraphics[scale=0.062]{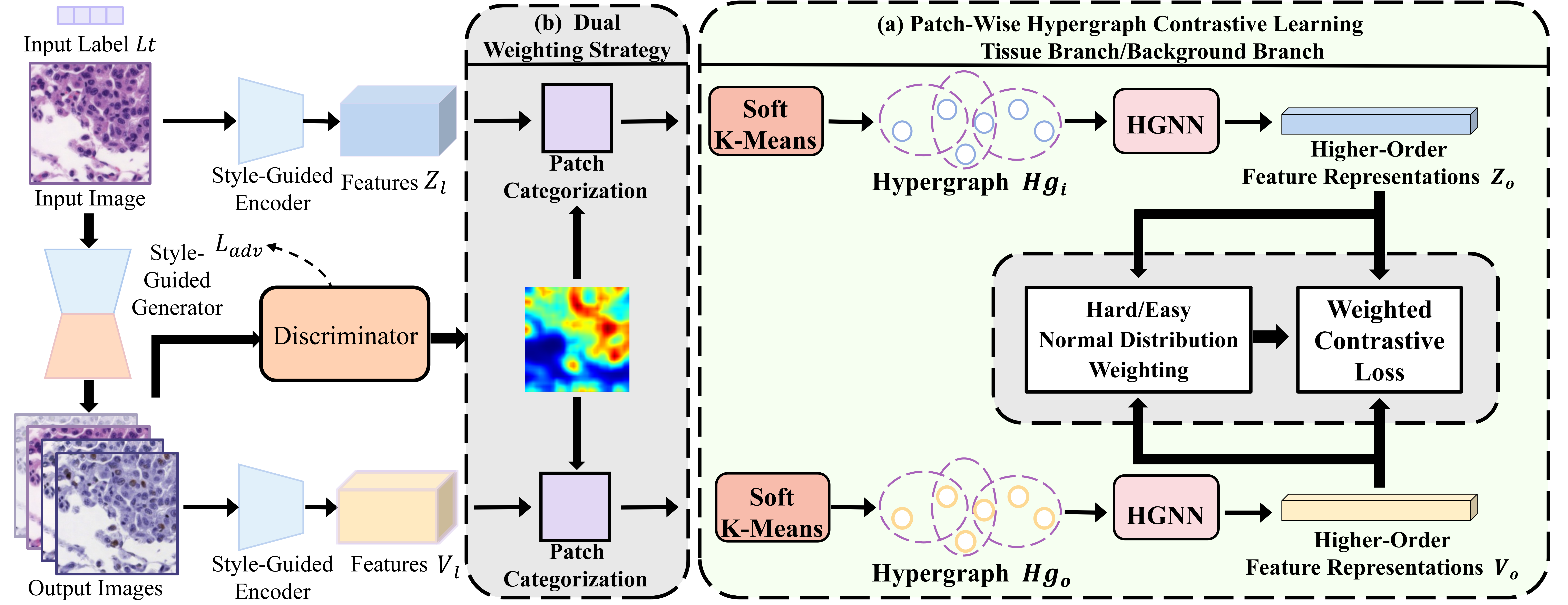}} 
\caption{Overall framework of the proposed method. (a) Our patch-wise hypergraph contrastive learning framework. We promote cross-patch higher-order topological consistency by maximizing the mutual information between the topological features of the input and output images. (b) A dual normal distribution weighting strategy is applied to negative samples to optimize the learning process, leveraging the distinct features of tissue and background regions. }
\label{fig:2}
\vspace{-15pt}
\end{figure*}

We use soft k-means clustering to group the $K$ patches for constructing the input branch hypergraph $Hg_i$ and output branch hypergraph $Hg_o$ (Fig. \ref{fig:2}(a)), which differs from the use of fuzzy c-means for hypergraph construction in \cite{han2023vision}. The clustering is applied to the patch features $Z_l^k \in \mathbb{R}^c$, producing a membership matrix $m$ and a centroid matrix $c$. The membership matrix $m$ represents the degree of membership of each patch to each hyperedge, while the centroid matrix $c$ indicates the central feature location of each hyperedge. For each hyperedge, we apply a pre-defined membership threshold to select the patches belonging to that hyperedge. Specifically, we identify the patches in the membership matrix $m$ whose membership values exceed the threshold, and store the indices of these patches in the hyperedge matrix $H_i$ and $H_o$, thus forming the final hyperedge structure. 

In image translation tasks, we seek to avoid spending excessive computational resources on mapping patch features $Z_l^k$ and $V_l^k$. Soft k-means offers computational advantages by enabling nodes to belong to multiple clusters simultaneously without the extensive complexity of fuzzy c-means. It achieves this by directly using Euclidean distance to perform soft assignments via a softmax function with temperature, which eliminates the need for complex operations like exponentiation and normalization across clusters.  Additionally, while fuzzy c-means is sensitive to the choice of the fuzziness parameter $m$, soft k-means exhibits lower complexity in parameter tuning, reducing the risk of model instability and minimizing the need for task-specific adjustments in stain transfer. 

After constructing the hypergraphs, we perform hypergraph convolution on the patch node feature sets as follows: 
\begin{equation} 
Z_o = D_e^{-\frac{1}{2}}  H_i \ \sigma \left( D_v^{-1} H_i^T \sigma \left( D_e^{-\frac{1}{2}} Z_i \Theta_{Z1} \right) \right) \Theta_{Z2} 
\end{equation} 
\begin{equation} 
V_o = D_e^{-\frac{1}{2}} H_o  \ \sigma \left( D_v^{-1} H_o^T \sigma \left( D_e^{-\frac{1}{2}} V_i \Theta_{V1} \right) \right) \Theta_{V2} 
\end{equation} 
where $\Theta_{Z1}, \Theta_{Z2}, \Theta_{V1}, \Theta_{V2}$ are learnable parameters of the HGNN layers, and $\sigma$ represents an activation function. $Z_i$ and $Z_o$ denote the input and output node embeddings of the input image branch, respectively, while $V_i$ and $V_o$ represent the input and output node embeddings of the output image branch. Hypergraph convolution can be understood as a two-step message-passing process, where information flows in a "node-hyperedge-node" manner. The multiplication with $H_i^T$ and $H_o^T$ achieves aggregation from nodes to hyperedges, while the multiplication with $H_i$ and $H_o$ can be seen as aggregating information from hyperedges back to nodes. $D_e$ and $D_v$ perform normalization. In summary, the HGNN layer effectively extracts high-order relationships within the hypergraph through node-edge-node transformations. 
Finally, we maximize the patch-wise mutual information between features $Z_o$ and $V_o$ from both branches: 
\begin{equation}
\mathcal{L}_{\text{STHCL}}(X, Y) = -\frac{1}{K} \sum_{i=1}^{K} \log \frac{e^{\frac{z_i^\top \cdot v_i }{\tau}}}{e^{\frac{z_i^\top \cdot v_i }{\tau}} + \sum_{\substack{j=1 \\ j \neq i}}^{K}e^{\frac{z_i^\top \cdot v_j }{\tau}}}
\end{equation}
where $z_i$ and $v_i$ are the $i$-th node features from $Z_o$ and $V_o$ and $\tau$ is the temperature parameter. 
\subsection{Dual Normal Distribution Weighting}
In MoNCE \cite{zhan2022modulated}, different weighting strategies are applied in paired and unpaired scenarios, with hard and easy weighting strategies applied to unpaired and paired image translation tasks, respectively. The hard-weighting strategy assigns higher weights to hard negative samples—those highly similar to the anchor, emphasizing their impact on the contrastive objective. Meanwhile, the easy-weighting strategy focuses on highlighting the contributions of moderately negative samples, and MoNCE achieves this by assigning lower weights to hard negative samples. MoNCE loss is defined as:
\begin{equation}
\mathcal{L}_{\text{MoNCE}}(X, Y) = 
-\frac{1}{K} \sum_{i=1}^{K} \log \frac{e^{\frac{z_i^\top \cdot v_i }{\tau}}}{e^{\frac{z_i^\top \cdot v_i }{\tau}} + \sum_{\substack{j=1 \\ j \neq i}}^{K}w_{ij}\cdot e^{\frac{z_i^\top \cdot v_j }{\tau}}}
\end{equation}

\begin{figure}[htbp]

\centerline{\includegraphics[scale=0.062]{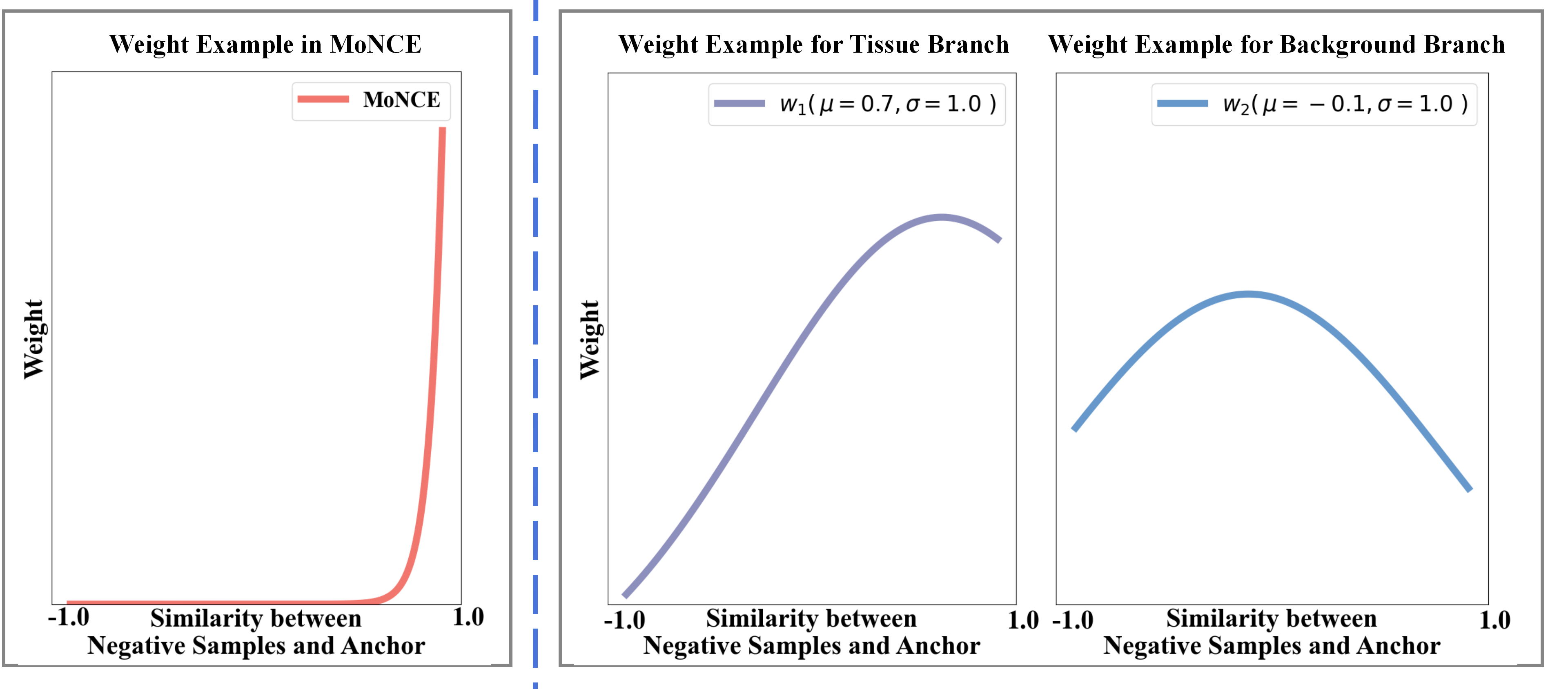}} 
\caption{Illustration of the relationship between the training weights assigned to negative samples and their similarity to the anchor point ($z_i^\top \cdot v_j$). The MoNCE approach places significant emphasis on hard negative samples, while our proposed method adopts a more balanced approach, incorporating both hard and easy negative sample weighting strategies depending on the context of the tissue and background regions.}
\label{fig3}
\vspace{-19pt}
\end{figure}

where $z_i$ and $v_j$ represent the feature embeddings of the patches, and $K$ is the total number of patches.

The hard-weighting strategy weight $w_{ij}^+$ and easy-weighting strategy weight $w_{ij}^-$ are defined as:
\begin{align}
w_{ij}^+ &= \frac{e^{\frac{z_i^\top v_j}{\tau}}}{\sum_{j=1}^{K} e^{\frac{z_i^\top v_j}{\tau}}},   
\ w_{ij}^- = \frac{e^{\frac{1 - z_i^\top v_j}{\tau}}}{\sum_{j=1}^{K} e^{\frac{1 - z_i^\top v_j}{\tau}}}
\end{align}

\begin{figure*}[htbp]
\centering
\includegraphics[width=\textwidth]{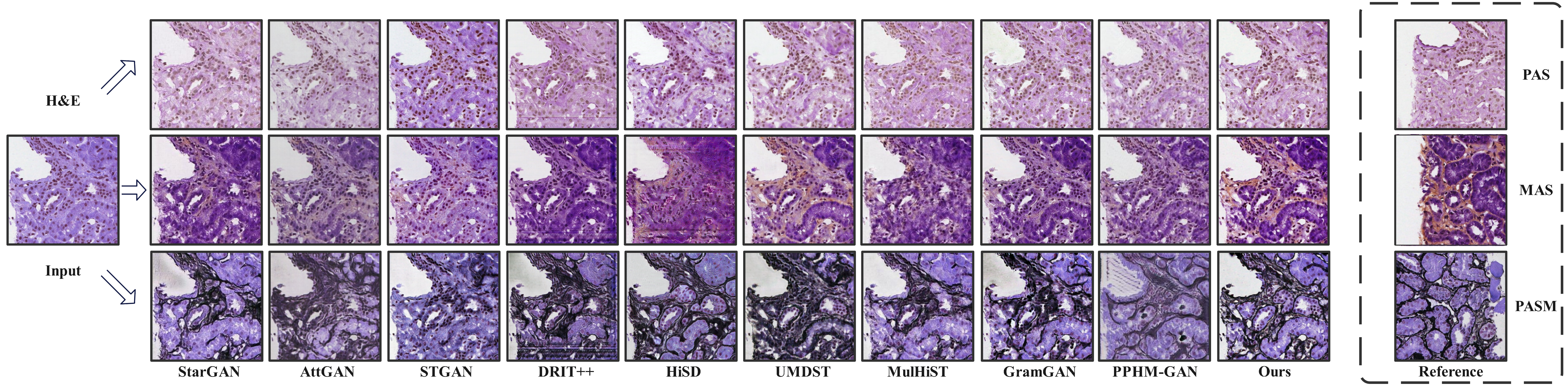} 
\caption{The performance comparison of various existing methods and our proposed method for multiple stain transfer of the same H\&E-stained image.
}
\label{fig4}
\vspace{-15pt}
\end{figure*}

Histopathological images often contain distinct tissue areas and white backgrounds. Previous works have generally applied the hard-weighting strategy to the entire image. However, we argue that using different strategies to the tissue and background regions maximizes the utility of negative samples. Specifically, in the background regions, different patches exhibit small differences in feature space, with overlapping similarity distributions between negative and positive samples. This situation is similar to paired image translation. Therefore, we adopt the easy-weighting strategy for background regions, assigning lower weights to the most similar negative samples to the anchor point. In contrast, we treat the tissue region’s stain transfer as a typical unpaired image translation task, applying the hard-weighting strategy with higher weights for negative samples closer to the anchor.

However, in MoNCE, weights are always exponentially related to the similarity between features. For the hard-weighting strategy, the sample weights are excessively high in high-similarity regions, while for the easy-weighting strategy, they may be too high in low-similarity regions. We consider this approach inflexible and too extreme. Additionally, in practice, for the hard-weighting strategy, the most informative negatives lie in the 'head' region (top 20\% similarity), not just the top 1\% \cite{cai2020all}, \cite{wu2024understanding}. The same holds for the easy-weighting strategy.
To address this, we introduce normal distribution weights into the image translation task, improving upon MoNCE's extremity. The normal distribution weight function allows for flexible weighting of negative samples. Example weights are shown in Fig. \ref{fig3}. We categorize tissue and background regions using the discriminator's heatmap output, without extra parameters \cite{cai2024rethinking}. This works because the tissue region is more complex in structure and diverse in morphology, causing the model to focus more on these areas, resulting in higher heatmap values. In contrast, the background's simpler color and structure make it easier for the model to distinguish real from fake, resulting in lower heatmap values. As shown in Fig. 2(b). Based on the heatmap values corresponding to the patches, we sample patches with high and low heatmap values to form two separate sets: hard features (high heatmap values) and easy features (low heatmap values). We then apply soft k-means clustering to each of the two separated feature sets to construct the hypergraphs, and perform hypergraph convolution on them. Finally, we apply different weight computation strategies to calculate the contrastive loss.

For the weighted loss function:
\begin{equation}
\mathcal{L}_{\text{w}}(X, Y) 
= -\frac{1}{K} \sum_{i=1}^{K}\log \frac{e^{\frac{z_i^\top \cdot v_i }{\tau}}}{e^{\frac{z_i^\top \cdot v_i }{\tau}} + \sum_{\substack{j=1
 \\ j \neq i}}^{K} w_{ij}(\mu, \sigma, \tau) e^{\frac{z_i^\top \cdot v_j }{\tau}}} 
\end{equation}
We define the weight as follows:
\begin{equation}
w_{ij}(\mu, \sigma, \tau) = \frac{\frac{1}{\sigma \sqrt{2\pi}} \exp \left( -\frac{(l_{\text{neg}}^{(ij)} - \mu)^2}{2\sigma^2} \right)}{\frac{1}{K} \sum_{m=1}^{K} \left( \frac{1}{\sigma \sqrt{2\pi}} \exp \left( -\frac{(l_{\text{neg}}^{(i,m)} - \mu)^2}{2\sigma^2} \right) \right)}
\end{equation}
where

\vspace{-2pt}
\begin{equation}
l_{\text{neg}}^{(ij)} = \frac{z_i^\top \cdot v_j}{\tau}
\end{equation}

\vspace{-2pt}
$\mu$ and $\sigma$ are controllable hyperparameters. $\mu$ controls the central region of weight distribution, with samples closer to $\mu$ having larger weights, while $\sigma$ controls the height of the weight distribution in the central region.
\vspace{-2pt} 
The overall contrastive learning loss function is:
\begin{align}
&\mathcal{L}_{\text{STNHCL}}(X, Y)  \nonumber
\\& = -\frac{1}{K} \sum_{i=1}^{K} \log \frac{e^{\frac{a_i^\top b_i}{\tau}}}{e^{\frac{a_i^\top b_i}{\tau}} + \sum_{\substack{j=1 \\ j \neq i}}^{K} w_{ij}(\mu_1, \sigma_1, \tau) e^{\frac{a_i^\top b_j}{\tau}}}  \nonumber \\
& \ \ \  -\frac{1}{K} \sum_{p=1}^{K} \log \frac{e^{\frac{c_p^\top d_p}{\tau}}}{e^{\frac{c_p^\top d_p}{\tau}} + \sum_{\substack{q=1 \\ q \neq p}}^{K} w_{pq}(\mu_2, \sigma_2, \tau) e^{\frac{c_p^\top d_q}{\tau}}}
\end{align}
Here, $a_i$ and $b_i$ represent the i-th patch features from the tissue parts of source and generated images, respectively, while $c_i$ and $d_i$ represent the i-th patch features from the background. The larger value of $\mu_1$ reflects the hard-weighting strategy, while smaller $\mu_2$ indicates easy weighting strategy.

The total loss of the generator can be defined as:
\begin{equation}
\mathcal{L}_{\text{total}} = \lambda_1 \times (\mathcal{L}_{\text{adv}} +\mathcal{L}_R + \mathcal{L}_{\text{non-dia}})\\+ \lambda_2\times(\mathcal{L}_{\text{STNHCL}}+\mathcal{L}_{\text{PatchNCE}})
\end{equation}
\vspace{-2pt} 
The adversarial loss can be defined as:
\begin{equation}
\begin{aligned}
\mathcal{L}_{\text{adv}} = & \mathbb{E}_{x \sim Y} \left[ \left( D(x) \right)^2 \right]+ \mathbb{E}_{x \sim X} \left[ \left( 1 - D(G(x, L_t)) \right)^2 \right]
\end{aligned}
\end{equation}
where $L_t$ is the target stain label, $G$ is the generator, and $D$ is the discriminator. In addition, $\mathcal{L}_R$ and $\mathcal{L}_{\text{non-dia}}$ are defined in \cite{guan2024unsupervised}, and $\mathcal{L}_{\text{PatchNCE}}$ is from CUT \cite{park2020contrastive}. Here, $\lambda_1 = \lambda_2 = 10$.
\vspace{-3pt} 
\section{Experiment}
\subsection{Datasets and Experiment Setup}\label{AA}
\vspace{-2pt} 
\textbf{Datasets.} We used three publicly available datasets: the latest versions of the human kidney dataset and human lung lesion dataset from ANHIR \cite{borovec2020anhir}, as well as Glomeruli Segmentation Dataset \cite{guan2024unsupervised} for downstream tasks. Details on data preprocessing can be found in the supplementary materials.


\textbf{Evaluation Metrics.} We employed standard metrics to evaluate the performance of pathological image translation: Contrast Structure Similarity (CSS), Fréchet Inception Distance (FID), and Kernel Inception Distance (KID). CSS measures how well the original structure and content of the image preserved. FID and KID are employed to quantify the distributional differences between real and generated images.

\textbf{Implementation Details.} Our method was implemented in PyTorch on an AMD EPYC 9754 CPU and NVIDIA RTX 3090 GPU. We used the generator from GramGAN \cite{guan2024unsupervised}. The training dataset patch size was set to $256 \times 256$, with a batch size of 1, and the model was trained for 300,000 iterations.
\vspace{-5pt}
\subsection{Comparison}
\vspace{-2pt}
We compared our method with state-of-the-art multi-domain stain transfer approaches and other applicable multi-domain image translation methods, including StarGAN \cite{choi2018stargan}, AttGAN \cite{he2019attgan}, STGAN \cite{liu2019stgan}, DRIT++ \cite{lee2020drit++}, HiSD \cite{li2021image}, UMDST  \cite{lin2022unpaired}, MulHiST \cite{shi2023mulhist}, GramGAN \cite{guan2024unsupervised}, and PPHM-GAN \cite{kawai2024virtual}. Quantitative results are shown in Table \ref{tab1}. Compared to baseline methods, our approach excelled in preserving the original tissue structure (measured by CSS) and generating realistic target stain styles (measured by FID and KID). In the human lung lesion dataset, STGAN achieved the best CSS metric but it performed poorly on FID and KID. Our approach demonstrated superior stain transfer performance. The visual qualitative results in Fig. \ref{fig4} show that our model generates clearer, more contrast-rich images with better structural and semantic alignment compared to other methods.


\vspace{-5pt} 
\begin{table}[ht]
\centering
\caption{Quantitative Results of Stain Transfer. The \textbf{bold} denotes the best and the \underline{underlined} denotes the second best.}
\label{tab1}
\begin{tabular}{@{}c c c c c@{}}
\toprule
Dataset & Method & CSS($\uparrow$) & FID($\downarrow$) & KID$\times100$($\downarrow$)\\
\midrule
\multirow{8}{*}{\begin{tabular}[c]{@{}c@{}}  \\ \\  Human Lung\\ Lesion Dataset \end{tabular}} & StarGAN & 0.823 & 60.30 & 3.41 \\
& AttGAN & 0.897 & 43.41 & \underline{2.68} \\
& STGAN & \underline{0.930} & 95.49 & 5.66 \\
& DRIT++ & 0.707 & 144.81 & 14.60 \\
& HiSD & 0.892 & 43.27 & 2.81 \\
& UMDST & 0.893 & \underline{40.14} & 2.77 \\
& MulHiST & 0.524 & 193.36 & 16.91 \\
& GramGAN & 0.875 & 56.96 & 2.73 \\
& PPHM-GAN & 0.856 & 64.21 & 3.42 \\
& \textbf{Ours} & \textbf{0.943} & \textbf{35.66} & \textbf{2.67} \\
\midrule
\multirow{8}{*}{\begin{tabular}[c]{@{}c@{}} \\ \\ Human Kidney \\ Dataset \end{tabular}} & StarGAN & 0.162 & 81.90 & \underline{7.78} \\
& AttGAN & 0.609 & 84.47 & 8.03 \\
& STGAN & \textbf{0.860} & 135.61 & 13.72 \\
& DRIT++ & 0.646 & 79.02 & 10.11 \\
& HiSD & 0.620 & 77.95 & 8.99 \\
& UMDST & 0.637 & \underline{77.91} & 8.94 \\
& MulHiST & 0.767 & 112.18 & 11.23 \\
& GramGAN & 0.657 & 88.46 & 8.80 \\
& PPHM-GAN & 0.464 & 171.07 & 16.82 \\
& \textbf{Ours} & \underline{0.776} & \textbf{56.53} & \textbf{5.75} \\
\bottomrule
\end{tabular}
\vspace{-13pt}
\end{table}
\begin{table}[htbp]
\caption{Quantitative results of ablation studies}
\label{tab2}
\centering
\begin{tabular}{ccc}
\toprule
Method & CSS($\uparrow$) & FID($\downarrow$) \\
\midrule
$\mathcal{L}_{\text{adv}}$ & 0.814 & 97.18 \\
$\mathcal{L}_{\text{adv}} + \mathcal{L}_{\text{cyc}}$ & 0.875 & 56.87 \\
$\mathcal{L}_{\text{adv}} + \mathcal{L}_{\text{patchNCE}}$ & 0.909 & 54.04 \\
$\mathcal{L}_{\text{adv}} + \mathcal{L}_{\text{PatchNCE}} + \mathcal{L}_{\text{PatchGCL}}$ & 0.917 & 46.69 \\
\textbf{$\mathcal{L}_{\text{adv}} + \mathcal{L}_{\text{patchNCE}} + \mathcal{L}_{\text{STHCL}}$ (\textbf{Ours})} & \textbf{0.928} & \textbf{41.53} \\
\midrule
$\mathcal{L}_{\text{adv}} + \mathcal{L}_{\text{patchNCE}} + \mathcal{L}_{\text{STHCL+traditional  
 weighting}}$ & 0.933 & 41.85 \\
\textbf{$\mathcal{L}_{\text{adv}} + \mathcal{L}_{\text{PatchNCE}} + \mathcal{L}_{\text{STNHCL}}$ (\textbf{Ours})} & \textbf{0.943} & \textbf{35.66} \\
\bottomrule
\end{tabular}
\vspace{-15pt}
\end{table}

\setlength{\tabcolsep}{3.8pt} 
\begin{table}[htbp]
  \centering
  \caption{Comparison of the Effects of Different Mean Pairs on CSS and FID Metrics on the Human Lung Lesion Dataset}
  \label{tab3}
  \begin{tabular}{c p{0.7cm} p{0.7cm} p{0.7cm} p{0.7cm} p{0.7cm} p{0.7cm} p{0.7cm} p{0.7cm}}
    \toprule
    \diagbox{$\mu_2$}{$\mu_1$} & \multicolumn{2}{c}{0.3} & \multicolumn{2}{c}{0.5} & \multicolumn{2}{c}{0.7} & \multicolumn{2}{c}{0.9} \\ 
    \cmidrule(lr){2-3} \cmidrule(lr){4-5} \cmidrule(lr){6-7} \cmidrule(lr){8-9}
    & \hspace{0.07cm}CSS & \hspace{0.1cm}FID & \hspace{0.07cm}CSS & \hspace{0.1cm}FID & \hspace{0.07cm}CSS & \hspace{0.1cm}FID & \hspace{0.07cm}CSS & \hspace{0.1cm}FID \\
    \midrule
    -0.2 & 0.935 & 43.07 & 0.936 & 41.76 & 0.936 & 39.29 & 0.936 & \textbf{35.66} \\
    0.1 & 0.940 & 41.79 & 0.940 & 38.43 & \textbf{0.943} & \textbf{35.66} & 0.943 & 40.40 \\
    0.3 & 0.939 & 40.56 & 0.940 & 39.16 & 0.942 & 38.02 & 0.935 & 41.64 \\
    0.5 & 0.937 & 40.14 & 0.940 & 39.54 & 0.941 & 41.48 & 0.934 & 42.37 \\
    0.7 & 0.936 & 40.32 & 0.939 & 42.58 & 0.941 & 41.73 & 0.932 & 42.20 \\
    \bottomrule
  \end{tabular} 
  \vspace{-12pt}
\end{table}

\begin{table}[h]
\centering
\caption{the quantitative results of the downstream tasks, we use map@[0.50:0.95] to measure the accuracy.}
\label{tab4}
\begin{tabular}{lcccc}
\toprule
Tasks       & \makecell{H\&E \\ (real)} & \makecell{PASM \\ (generated)} & \makecell{PAS \\ (generated)}   & \makecell{MAS \\ (generated)}  \\ \midrule
Detection             & 0.380               & \textbf{0.446}                    & 0.408          & 0.406         \\
Segmentation          & 0.402               & \textbf{0.461}                    & 0.386          & 0.389         \\ \bottomrule
\end{tabular}
\vspace{-13pt}
\end{table}
\subsection{Ablations}
\vspace{-2pt}
We conducted ablation studies to assess the effectiveness of the proposed method. These include experiments on PatchNCE, STHCL, and STNHCL. The experimental results are shown in Table \ref{tab2}. The results indicate that each component significantly enhances the model's performance. Moreover, our $\mathcal{L}_{\text{STHCL}}$ outperforms $\mathcal{L}_{\text{PatchGCL}}$ in enhancing the higher-order semantic correspondence across multiple patches. Our dual normal distribution weighting strategy leverages negative samples more effectively compared to traditional weighting strategies. Furthermore, we evaluated the performance of STNHCL under different settings of mean parameters, as shown in Table \ref{tab3}. The $\mu_1$ and $\mu_2$ represent the normal weighting means for the tissue region and the background region, respectively. We observed that the best performance was achieved when the mean was set at an intermediate value, rather than at the extremes, confirming the effectiveness of our dual normal distribution weighting mechanism.
\vspace{-10pt} 
\subsection{Downstream Tasks}
On the Glomeruli Segmentation Dataset \cite{guan2024unsupervised}, we utilize the same network (Mask R-CNN \cite{he2017mask}) to detect and segment glomeruli in both H\&E-stained images and their virtually converted stained counterparts generated by our model. The virtually generated stains achieve performance that is comparable to or even surpasses the original H\&E stains in detection and segmentation tasks, as demonstrated in Table \ref{tab4}.
\vspace{-4.49pt} 
\section{Conclusion}
\vspace{-2.5pt} 
Traditional stain transfer methods often struggle to maintain higher-order semantic correspondence between multiple patches of input and output images, leading to the loss of essential information. To address this issue, we introduce STNHCL, which constructs hypergraph-based feature representations to match the input and output images. To further improve upon the traditional extreme negative sample weighting strategy, which does not fully utilize effective negative sample information, we employ a more moderate normal distribution weighting scheme. Experimental results demonstrate superior performance, further validating the advantages of our method.

\vspace{-3.2pt} 
\nocite{*}  
\typeout{=== FORCE BIBTEX ===}  
\bibliographystyle{IEEEbib}
\bibliography{icme2025references}

\begin{thebibliography}{10}

\bibitem{bai2023deep}
Bijie Bai, Xilin Yang, Yuzhu Li, Yijie Zhang, Nir Pillar, and Aydogan Ozcan,
\newblock ``Deep learning-enabled virtual histological staining of biological samples,''
\newblock {\em Light: Science \& Applications}, vol. 12, no. 1, pp. 57, 2023.

\bibitem{he2017mask}
Kaiming He, Georgia Gkioxari, Piotr Doll{\'a}r, and Ross Girshick,
\newblock ``Mask r-cnn,''
\newblock in {\em Proceedings of the IEEE international conference on computer vision}, 2017, pp. 2961--2969.

\bibitem{lo2021cycle}
Ying-Chih Lo, I-Fang Chung, Shin-Ning Guo, Mei-Chin Wen, and Chia-Feng Juang,
\newblock ``Cycle-consistent gan-based stain translation of renal pathology images with glomerulus detection application,''
\newblock {\em Applied Soft Computing}, vol. 98, pp. 106822, 2021.

\bibitem{lin2022unpaired}
Yiyang Lin, Bowei Zeng, et~al.,
\newblock ``Unpaired multi-domain stain transfer for kidney histopathological images,''
\newblock in {\em Proceedings of the AAAI Conference on Artificial Intelligence}, 2022, vol.~36, pp. 1630--1637.

\bibitem{kawai2024virtual}
Masataka Kawai, Toru Odate, Kazunari Kasai, Tomohiro Inoue, Kunio Mochizuki, et~al.,
\newblock ``Virtual multi-staining in a single-section view for renal pathology using generative adversarial networks,''
\newblock {\em Computers in Biology and Medicine}, vol. 182, pp. 109149, 2024.

\bibitem{guan2024unsupervised}
Xianchao Guan, Yifeng Wang, Yiyang Lin, Xi~Li, and Yongbing Zhang,
\newblock ``Unsupervised multi-domain progressive stain transfer guided by style encoding dictionary,''
\newblock {\em IEEE Transactions on Image Processing}, 2024.

\bibitem{zhu2017unpaired}
Jun-Yan Zhu, Taesung Park, Phillip Isola, and Alexei~A Efros,
\newblock ``Unpaired image-to-image translation using cycle-consistent adversarial networks,''
\newblock in {\em Proceedings of the IEEE international conference on computer vision}, 2017, pp. 2223--2232.

\bibitem{park2020contrastive}
Taesung Park, Alexei~A Efros, Richard Zhang, and Jun-Yan Zhu,
\newblock ``Contrastive learning for unpaired image-to-image translation,''
\newblock in {\em Computer Vision--ECCV 2020: 16th European Conference, Glasgow, UK, August 23--28, 2020, Proceedings, Part IX 16}. Springer, 2020, pp. 319--345.

\bibitem{zhan2022modulated}
Fangneng Zhan, Jiahui Zhang, Yingchen Yu, Rongliang Wu, and Shijian Lu,
\newblock ``Modulated contrast for versatile image synthesis,''
\newblock in {\em Proceedings of the IEEE/CVF Conference on Computer Vision and Pattern Recognition}, 2022, pp. 18280--18290.

\bibitem{hu2022qs}
Xueqi Hu, Xinyue Zhou, Qiusheng Huang, Zhengyi Shi, Li~Sun, and Qingli Li,
\newblock ``Qs-attn: Query-selected attention for contrastive learning in i2i translation,''
\newblock in {\em Proceedings of the IEEE/CVF Conference on Computer Vision and Pattern Recognition}, 2022, pp. 18291--18300.

\bibitem{jung2024patch}
Chanyong Jung, Gihyun Kwon, and Jong~Chul Ye,
\newblock ``Patch-wise graph contrastive learning for image translation,''
\newblock in {\em Proceedings of the AAAI Conference on Artificial Intelligence}, 2024, vol.~38, pp. 13013--13021.

\bibitem{han2023vision}
Yan Han, Peihao Wang, Souvik Kundu, Ying Ding, and Zhangyang Wang,
\newblock ``Vision hgnn: An image is more than a graph of nodes,''
\newblock in {\em Proceedings of the IEEE/CVF International Conference on Computer Vision}, 2023, pp. 19878--19888.

\bibitem{gao20123}
Yue Gao, Meng Wang, Dacheng Tao, Rongrong Ji, and Qionghai Dai,
\newblock ``3-d object retrieval and recognition with hypergraph analysis,''
\newblock {\em IEEE transactions on image processing}, vol. 21, no. 9, pp. 4290--4303, 2012.

\bibitem{huang2009video}
Yuchi Huang, Qingshan Liu, and Dimitris Metaxas,
\newblock ``] video object segmentation by hypergraph cut,''
\newblock in {\em 2009 IEEE conference on computer vision and pattern recognition}. IEEE, 2009, pp. 1738--1745.

\bibitem{zhou2022hypergraph}
Yuxuan Zhou, Zhi-Qi Cheng, Chao Li, Yanwen Fang, Yifeng Geng, et~al.,
\newblock ``Hypergraph transformer for skeleton-based action recognition,''
\newblock {\em arXiv preprint arXiv:2211.09590}, 2022.

\bibitem{huang2023reconstructing}
Buzhen Huang, Jingyi Ju, Zhihao Li, and Yangang Wang,
\newblock ``Reconstructing groups of people with hypergraph relational reasoning,''
\newblock in {\em Proceedings of the IEEE/CVF International Conference on Computer Vision}, 2023, pp. 14873--14883.

\bibitem{cai2020all}
Tiffany~Tianhui Cai, Jonathan Frankle, David~J Schwab, and Ari~S Morcos,
\newblock ``Are all negatives created equal in contrastive instance discrimination?,''
\newblock {\em arXiv preprint arXiv:2010.06682}, 2020.

\bibitem{wu2024understanding}
Junkang Wu, Jiawei Chen, Jiancan Wu, et~al.,
\newblock ``Understanding contrastive learning via distributionally robust optimization,''
\newblock {\em Advances in Neural Information Processing Systems}, vol. 36, 2024.

\bibitem{cai2024rethinking}
Xiuding Cai, Yaoyao Zhu, Dong Miao, Linjie Fu, and Yu~Yao,
\newblock ``Rethinking the paradigm of content constraints in unpaired image-to-image translation,''
\newblock in {\em Proceedings of the AAAI Conference on Artificial Intelligence}, 2024, vol.~38, pp. 891--899.

\bibitem{borovec2020anhir}
Ji{\v{r}}{\'\i} Borovec, Jan Kybic, Ignacio Arganda-Carreras, et~al.,
\newblock ``Anhir: automatic non-rigid histological image registration challenge,''
\newblock {\em IEEE transactions on medical imaging}, vol. 39, no. 10, pp. 3042--3052, 2020.

\bibitem{choi2018stargan}
Yunjey Choi, Minje Choi, Munyoung Kim, Jung-Woo Ha, et~al.,
\newblock ``Stargan: Unified generative adversarial networks for multi-domain image-to-image translation,''
\newblock in {\em Proceedings of the IEEE conference on computer vision and pattern recognition}, 2018, pp. 8789--8797.

\bibitem{he2019attgan}
Zhenliang He, Wangmeng Zuo, Meina Kan, Shiguang Shan, and Xilin Chen,
\newblock ``Attgan: Facial attribute editing by only changing what you want,''
\newblock {\em IEEE transactions on image processing}, vol. 28, no. 11, pp. 5464--5478, 2019.

\bibitem{liu2019stgan}
Ming Liu, Yukang Ding, Min Xia, Xiao Liu, Errui Ding, et~al.,
\newblock ``Stgan: A unified selective transfer network for arbitrary image attribute editing,''
\newblock in {\em Proceedings of the IEEE/CVF conference on computer vision and pattern recognition}, 2019, pp. 3673--3682.

\bibitem{lee2020drit++}
Hsin-Ying Lee, Hung-Yu Tseng, Qi~Mao, et~al.,
\newblock ``Drit++: Diverse image-to-image translation via disentangled representations,''
\newblock {\em International Journal of Computer Vision}, vol. 128, no. 10-11, pp. 2402--2417, 2020.

\bibitem{li2021image}
Xinyang Li, Shengchuan Zhang, Jie Hu, Liujuan Cao, Xiaopeng Hong, Xudong Mao, et~al.,
\newblock ``Image-to-image translation via hierarchical style disentanglement,''
\newblock in {\em Proceedings of the IEEE/CVF conference on computer vision and pattern recognition}, 2021, pp. 8639--8648.

\bibitem{shi2023mulhist}
Lulin Shi, Yan Zhang, Ivy~HM Wong, Claudia~TK Lo, and Terence~TW Wong,
\newblock ``Mulhist: Multiple histological staining for thick biological samples via unsupervised image-to-image translation,''
\newblock in {\em International Conference on Medical Image Computing and Computer-Assisted Intervention}. Springer, 2023, pp. 735--744.

\end{thebibliography}

\end{document}